\documentclass[10pt,journal,compsoc]{IEEEtran}
\ifCLASSOPTIONcompsoc
  \usepackage[nocompress]{cite}
\else
  \usepackage{cite}
\fi
\usepackage{amssymb}

\ifCLASSINFOpdf
   \usepackage[pdftex]{graphicx}
\else
\fi
\usepackage{amsmath}
\usepackage{gensymb}
\usepackage{colortbl}

\usepackage{multirow}
\usepackage{acronym}
\newacro{CNN}[CNN]{Convolutional Neural Network}
\newacro{CCM}[CCM]{Color Conversion Matrix}
\newacro{RAE}[RAE]{recovery angular error}
\newacro{C3AE}[C3AE]{Color Constancy Convolutional AutoEncoder}
\newacro{HVS}[HVS]{Human Visual System}
\newacro{NLP}[NLP]{Natural Language Processing}
\newacro{GAN}[GAN]{Generative Adversarial Network}
\newacro{BoCF}[BoCF]{Bag of Color Features}
\newacro{ReLU}[ReLU]{Rectified Linear Units}
\makeatletter
\renewcommand\@makefntext[1]{\leftskip=2em\hskip-2em\@makefnmark#1}
\makeatother

\begin{document}
%
% paper title
% Titles are generally capitalized except for words such as a, an, and, as,
% at, but, by, for, in, nor, of, on, or, the, to and up, which are usually
% not capitalized unless they are the first or last word of the title.
% Linebreaks \\ can be used within to get better formatting as desired.
% Do not put math or special symbols in the title.
\title{Bag of Color Features For Color Constancy}

% author names and IEEE memberships
% note positions of commas and nonbreaking spaces ( ~ ) LaTeX will not break
% a structure at a ~ so this keeps an author's name from being broken across
% two lines.
% use \thanks{} to gain access to the first footnote area
% a separate \thanks must be used for each paragraph as LaTeX2e's \thanks
% was not built to handle multiple paragraphs
%
%
%\IEEEcompsocitemizethanks is a special \thanks that produces the bulleted
% lists the Computer Society journals use for "first footnote" author
% affiliations. Use \IEEEcompsocthanksitem which works much like \item
% for each affiliation group. When not in compsoc mode,
% \IEEEcompsocitemizethanks becomes like \thanks and
% \IEEEcompsocthanksitem becomes a line break with idention. This
% facilitates dual compilation, although admittedly the differences in the
% desired content of \author between the different types of papers makes a
% one-size-fits-all approach a daunting prospect. For instance, compsoc 
% journal papers have the author affiliations above the "Manuscript
% received ..."  text while in non-compsoc journals this is reversed. Sigh.

\author{Firas~Laakom, Nikolaos~Passalis,
Jenni~Raitoharju,~\IEEEmembership{Member,~IEEE,}
Jarno~Nikkanen,~\IEEEmembership{Member,~IEEE,}
Anastasios~Tefas,~\IEEEmembership{Member,~IEEE,}
Alexandros~Iosifidis,~\IEEEmembership{Senior~Member,~IEEE,} and~Moncef~Gabbouj,~\IEEEmembership{Fellow,~IEEE}

    %   \thanks{This work was supported by the NSF-Business Finland Center for Visual and Decision Informatics project Co-Botics, jointly sponsored by Tieto Oy Finland and CA Technologies (Broadcom recently acquired CA Technologies).}

       \thanks{F. Laakom, N. Passalis, J. Raitoharju,  and M. Gabbouj are with Faculty of Information Technology and Communication Sciences, Tampere University, Tampere, Finland (e-mail: firas.laakom@tuni.fi; nikolaos.passalis@tuni.fi; jenni.raitoharju@tuni.fi; moncef.gabbouj@tuni.fi).}
       \thanks{A. Iosifidis is with the Department of Engineering, Electrical and Computer Engineering, Aarhus University, DK-8200 Aarhus, Denmark (e-mail: ai@eng.au.dk).}
       
       \thanks{A. Tefas is with the Department of Informatics, Aristotle    University
        of Thessaloniki, 54124 Thessaloniki, Greece (e-mail: tefas@aiia.csd.auth.gr)}
       
       \thanks{J. Nikkanen is with INTEL, Insin\"{o}\"{o}rinkatu 41, 33720 Tampere, Finland (email: jarno.nikkanen@intel.com) }
%{\{fahadsohrab,haerdogan\}}@sabanciuniv.edu}
}
\IEEEtitleabstractindextext{%
\begin{abstract}
In this paper, we propose a novel color constancy approach, called \acf{BoCF},
building upon Bag-of-Features pooling. The proposed method substantially reduces the number of parameters needed for illumination estimation. At the same time, the proposed method is consistent  with the color constancy assumption stating that global spatial information is not relevant for illumination estimation and local information (edges, etc.) is sufficient. Furthermore, \ac{BoCF} is consistent with color constancy statistical approaches and can be interpreted as a learning-based generalization of many statistical approaches. To further improve the illumination estimation accuracy, we propose a novel attention mechanism for the \ac{BoCF} model with two variants based on self-attention. \ac{BoCF} approach and its variants achieve competitive, compared  to the state of the art, results while  requiring much fewer parameters on three benchmark datasets: ColorChecker RECommended, INTEL-TUT version 2, and NUS8. 
\end{abstract}

% Note that keywords are not normally used for peerreview papers.
}

% make the title area
\maketitle

% To allow for easy dual compilation without having to reenter the
% abstract/keywords data, the \IEEEtitleabstractindextext text will
% not be used in maketitle, but will appear (i.e., to be "transported")
% here as \IEEEdisplaynontitleabstractindextext when the compsoc 
% or transmag modes are not selected <OR> if conference mode is selected 
% - because all conference papers position the abstract like regular
% papers do.
\IEEEdisplaynontitleabstractindextext
% \IEEEdisplaynontitleabstractindextext has no effect when using
% compsoc or transmag under a non-conference mode.

% For peer review papers, you can put extra information on the cover
% page as needed:
% \ifCLASSOPTIONpeerreview
% \begin{center} \bfseries EDICS Category: 3-BBND \end{center}
% \fi
%
% For peerreview papers, this IEEEtran command inserts a page break and
% creates the second title. It will be ignored for other modes.
\IEEEpeerreviewmaketitle

\IEEEraisesectionheading{\section{Introduction}\label{sec:introduction}}
% Computer Society journal (but not conference!) papers do something unusual
% with the very first section heading (almost always called "Introduction").
% They place it ABOVE the main text! IEEEtran.cls does not automatically do
% this for you, but you can achieve this effect with the provided
% \IEEEraisesectionheading{} command. Note the need to keep any \label that
% is to refer to the section immediately after \section in the above as
% \IEEEraisesectionheading puts \section within a raised box.

% The very first letter is a 2 line initial drop letter followed
% by the rest of the first word in caps (small caps for compsoc).
% 
% form to use if the first word consists of a single letter:
% \IEEEPARstart{A}{demo} file is ....
% 
% form to use if you need the single drop letter followed by
% normal text (unknown if ever used by the IEEE):
% \IEEEPARstart{A}{}demo file is ....
% 
% Some journals put the first two words in caps:
% \IEEEPARstart{T}{his demo} file is ....
% 
% Here we have the typical use of a "T" for an initial drop letter
% and "HIS" in caps to complete the first word.

\IEEEPARstart{C}{olor} constancy in general is the ability of an imaging  system to discount the effects of illumination on the observed colors in a scene \cite{Ebner,cc}. When a person stands in a room lit by a colorful light, the \acf{HVS} unconsciously removes the lightening effects and the colors are perceived as if they were illuminated by a neutral, white light. While this ability is very natural for the \ac{HVS}, mimicking the same ability in a computer vision system is a challenging and under-constrained problem. Given a green pixel, one can not assert if it is a green pixel under a white illumination or a white pixel lit with a greenish illumination. Nonetheless, illumination estimation is considered an important component of many higher level computer vision tasks such as object recognition and tracking. Thus, it has been extensively studied in order to develop reliable color constancy systems which can achieve illumination invariance to some extent \cite{Ebner,limbert}.

The RGB image value $ \rho(x,y)$ in  the position $(x,y)$ of an image can be expressed as a function  depending on three  key factors \cite{limbert}: the illuminant distribution $ I(x,y, \lambda)$, the surface reflectance $R(x,y,\lambda)$ and the camera sensitivity $S(\lambda)$, where $\lambda$ is the wave length. This dependency is expressed as
\begin{equation}
            \rho(x,y) =  \int_\lambda I(x,y, \lambda) R(x,y,\lambda) S(\lambda) d \lambda. 
\end{equation}
Color constancy methods \cite{limbert,3} aim to estimate a uniform projection of $I( x,y,\lambda)$ on the sensor spectral sensitivities $S(\lambda)$, i.e.,
\begin{equation}
       I = I(x,y) =  \int_\lambda I(x,y, \lambda) S(\lambda) d \lambda,
\end{equation}
where $I$ is the global illumination of the scene.

\begin{figure}[t]
\centering
\includegraphics[width=8.5cm]{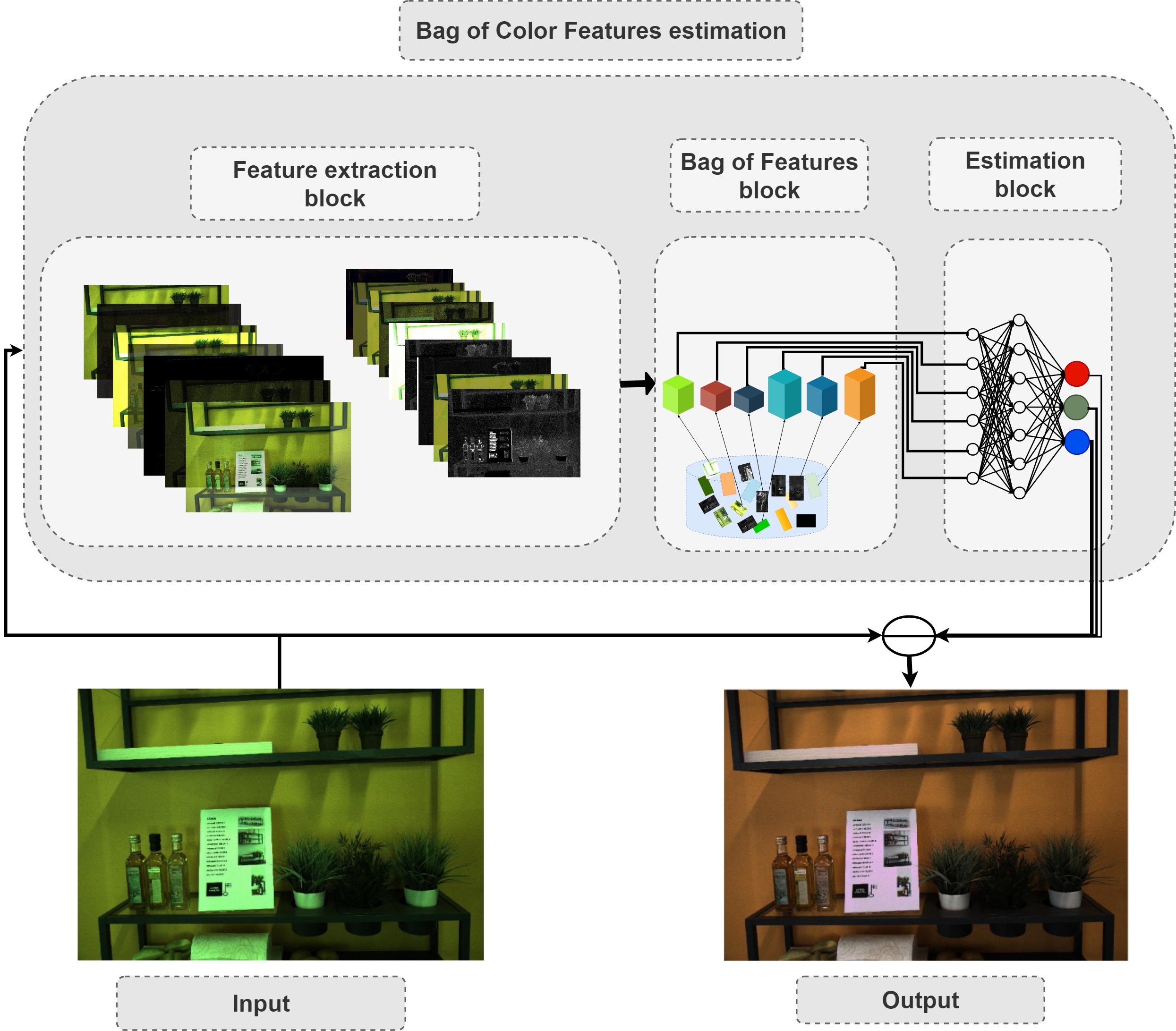}
% where an .eps filename suffix will be assumed under latex, 
% and a .pdf suffix will be assumed for pdflatex; or what has been declared
% via \DeclareGraphicsExtensions.
\caption{Building blocks of  \acf{BoCF} approach for illumination estimation}
\label{fig_1}
\end{figure}

Recently, deep learning approaches and \acp{CNN} in particular have become dominant in almost all computer vision tasks, including color constancy \cite{22,44,DSN,HoldGeoffroy2017DeepOI} due to their ability to take raw images directly as input and incorporate feature extraction in the learning process \cite{NIPS2012_4824}. Despite their accuracy in estimating illumination across multiple datasets \cite{17,47,44},  deploying \ac{CNN}-based approaches on low computational power devices, e.g., mobile devices, is still limited, since most of the high-accuracy deep models are computationally expensive \cite{44,DSN,HoldGeoffroy2017DeepOI}, which make them inefficient in terms of time and energy consumption. Additionally, most of the available datasets for illumination estimation are rather small-scale  \cite{Hemrit2018RehabilitatingTC,NUS,17} and hence not suitable for training large models. For this purpose, many state of the art approaches \cite{22,44} rely on pre-trained  networks to overcome this limitation. On the other hand, these pre-trained networks \cite{squueze,NIPS2012_4824} are  originally trained  for  a  classification  task. Thus, they are usually agnostic to the illumination color. This makes their usage in color constancy counter-intuitive as the illumination information is distorted in the early pre-trained layers. An alternative approach is of course to reduce the number of model parameters in order to use existing datasets, as shallower models, in general, need less examples to learn.

Furthermore, in \cite{NUS,8} it is argued that global spatial information is not an important feature in color constancy. The local information, i.e., the color distribution and the color gradient distribution (i.e. edges) can be sufficient  to extract the illumination information \cite{NUS}. Thus, using regular neural networks configurations to extract deep features is counter-intuitive in this particular problem. To address these drawbacks and challenges, we propose in this paper a novel color constancy deep learning approach called \acf{BoCF}. \ac{BoCF} uses Bag-of-Features Pooling\cite{bof}, which takes advantage of the ability of \acp{CNN} to learn relevant shallow features while  keeping the model suitable for low-power hardware. Furthermore, the proposed approach is consistent with the assumption that global spatial information is not relevant \cite{NUS,8} for color illumination estimation. 

Bag-of-Features Pooling is a neural extension \cite{bofp,bof } of the famous  Bag-of-Features model (BoF), also known as Bag-of-Visual Words (BoVW)\cite{boforigi , boforigi2}. BoFs  are wildly used in computer vision tasks, such as action recognition \cite{bof3}, object detection/recognition, sequence classification \cite{added}, and information retrieval \cite{bof2}. A BoF layer can be combined with convolutional layers to form a powerful convolutional architecture that is end-to-end trainable using the regular back-propagation algorithm \cite{bofp}.  

 The block diagram of the proposed \ac{BoCF} model is illustrated in Figure \ref{fig_1}. It consists of three main blocks: feature extraction block, Bag of Features block, and an estimation block. In the first block, regular convolutional layers are used to extract relevant features. Inspired by the assumption that second order gradient information is sufficient to extract the illumination information \cite{NUS}, we use only two convolutional layers to extract the features. In our experiments, we also study and validate this hypothesis empirically.  In the second block, i.e., the Bag of Features block, the network learns the dictionary using back-propagation\cite{bofp} over the non-linear transformation provided by the first block. This block outputs a histogram representation, which is fed to the last component, i.e., the estimation block, to regress to the scene illumination. 
 
 In most \ac{CNN}-based approaches used to solve the color constancy problem \cite{22,44,DSN,HoldGeoffroy2017DeepOI}, fully connected layers are connected directly to  a flattened version of the last convolutional layer output. This increases the numbers of parameters dramatically, as  convolutional layer outputs usually have a high dimensionality. In the proposed method, we address this problem by introducing an intermediate pooling block, i.e., the Bag of Features block, between the last convolutional layer and the fully connected layers. The proposed model achieves comparable results to previous state of the art illumination estimation methods while substantially reducing the number of the needed parameters, by up to 95\%. Additionally, the pooling process natively discards all global spatial information, which is, as discussed earlier, irrelevant for color constancy. Using only two convolutional layers in the first block, limits the model to only shallow features. These two advantages make proposed approach both consistent and in full corroboration with statistical approaches \cite{NUS}. 

To further improve the performance of the proposed model, we also propose two variants of a self-attention mechanism for the  \ac{BoCF} model. In the first variant, we add an attention mechanism between the feature extraction block and the Bag of Features block. This mechanism allows the network to dynamically select parts of the image to use for estimating the illumination, while discarding the remaining parts. Thus, the network becomes robust to noise and irrelevant features. In the second variant, we add an attention mechanism on top of the histogram representation, i.e.,  between the Bag of Features block and the estimation block. In this way, we allow the network to learn to adaptively select the elements of the histogram which best encode the illuminant information. The model looks over the whole histogram after the spatial information has been discarded and generates a proper representation according the current context (histogram). The introduced dynamics will be shown in the experiments to enhance the model performance with respect to all evaluation metrics and across all the datasets.

The main contributionsof the paper are as follows:
\begin{itemize}
\item We propose a novel \ac{CNN}-based color constancy algorithm, called \ac{BoCF}, based on Bag-of-Features Pooling. The proposed model is both shallow and able to achieve competitive results across multiple datasets compared to the state of the art.
\item We establish explicit links between \ac{BoCF} and prior statistical methods for illumination estimation and show that the proposed method can be framed as a learning-based generalization of many statistical approaches. This powerful approach fills the gap and provides the missing links between \ac{CNN}-based approaches and static approaches. 
\item We propose two novel attention mechanisms for \ac{BoCF} that can further improve the results. To the best of our knowledge, this is the first work which combines attention mechanism with Bag-of-Features Pooling.

\item The proposed method is extensively evaluated over three datasets leading to competitive performance with respect to existing state of the art, while substantially reducing the number of parameters. 
\end{itemize}
The rest of this paper is organized as follows. Section \ref{sec:related}
provides the background of color constancy approaches and a brief review of the Bag-of-Features Pooling technique and the attention mechanism used in this work. Section  \ref{sec:Proposed} details the proposed approach along with the two attention mechanisms based variants. Section \ref{sec:setup} introduces the datasets and the evaluation metrics used in this work along with the evaluation procedure. Section \ref{sec:res} presents the experimental results on three datasets: ColorChecker RECommended \cite{Hemrit2018RehabilitatingTC}, NUS8-Dataset\cite{NUS}, and INTEL-TUT version2\cite{17}. In Section  \ref{subsec:des}, we highlight the links between our approach and many existing methods and we show how our approach can be considered  as a generic framework for expressing existing approaches. Section \ref{sec:conc} concludes the paper.

\section{Related work}
\label{sec:related}
\subsection{Color constancy}
Typically, two types of color constancy approaches are distinguished, namely static methods and  supervised methods.  The former involves methods with static parameters settings that do not need any labeled image data for learning the model, while the latter are  data-driven approaches that learn to estimate the illuminant in a supervised manner using labeled data.

\subsubsection{Static methods}
Static methods exploit the statistical or physical properties of a scene by making assumptions about the nature of colors. They can be classified into two categories: methods based on low-level statistics\cite{2,4,5,ratina} and methods based on the physics-based dichromatic reflection model \cite{3,11,8,Category}.
 A number of approaches belonging to the first category were unified by Van de Weijer et al. \cite{5} into a single framework expressed as follows: 
\begin{equation}
                           \rho^{gt}(n,p, \sigma) =  \frac{1}{k}  (\int_x \int_y | \bigtriangledown^n  \rho_{\sigma}(x,y) |^p dxdy)^{ \frac{1}{p}},
\end{equation}
 where $n$  denotes the derivative order, $p$ the Minkowski norm and $k$  the normalization constant for $\rho^{gt}$. Also, $\rho_{\sigma}(x,y) = \rho(x,y) * g_{\sigma}(x,y) $ denotes the image convolution with a Gaussian filter with a scale parameter $\sigma$.
 This framework allows for deriving different algorithms simply by setting the appropriate values for $n$, $p$ and $\sigma$. The well-known Gray-World method \cite{4}, corresponding to $(n = 0,p =1, \sigma = 0)$,  assumes that  under a  neutral illumination the average reflectance in a scene is achromatic and the  illumination is estimated  as the shift of the image average color from gray. White-Patch \cite{2} $(n = 0,p = \infty, \sigma = 0)$, assumes that the maximum  values of RGB  color  channels are  caused by  a  perfectly reflecting  surface in the scene. Therefore, the illumination components correspond to these maximum values. Besides Gray-World and White-Patch methods, which make use of the color distribution  in the scene to  build their estimations, Gray-Edge method \cite{5} utilizes image  derivatives. Instead of the global average color, Gray-Edge methods $(n = 1,p = p, \sigma =\sigma  )$ assume that the average color of edges or the gradient of edges is gray. The illuminant's color is then estimated as the shift of the average edge color from gray.
 
Physics-based dichromatic  reflection models estimate the illumination by analyzing the scene and exploiting the physical interactions between the objects and the illumination. The main assumption of most methods in this category is that all pixels of a surface  form a plane in RGB color space. As the scene contains multiple surfaces, this results in multiple planes. The intersection between these planes is used to compute the color of the light source \cite{11}. Lee et al. \cite{8} exploited the bright areas in the captured scene to obtain an estimate of the illuminant color. In this work, we establish links between our proposed approach,  \ac{BoCF}, and several static methods. We show that  \ac{BoCF} can be interpreted as a learning-based extension of several of  these approaches.

\subsubsection{Supervised methods}

Supervised methods can be further divided into two main categories: characterization-based methods \cite{64,644} and training-based methods\cite{34,4664624,22,44}. The former involves ’light’ training processes in order to learn the characterization of the camera response in some way, while the latter involves methods that try to learn the illumination directly from the scene. 

Gamut Mapping \cite{64,644} is one of the most famous characterization-based approaches. It assumes that for a given illumination condition, only a limited number of colors can be observed. Thus any unexpected variation in the observed colors is caused by the light source illuminant. The set of colors that can occur under a given illumination, called canonical gamut, is first learned in a supervised manner. In the evaluation, an input gamut  which  represents  the  set  of  colors  used  to  acquire  the scene is constructed. The illumination is then estimated by mapping this input gamut to the canonical gamut.

Another group of training-based methods combines different illumination estimation approaches and learns a model that uses the best performing method or a combination of methods to estimate the illuminant of each input based on the scene characteristics \cite{34}. Bianco et al. used indoor/outdoor classification to select the optimal color constancy algorithm given an input image\cite{4664624}. Lu et al. proposed an approach which exploits 3D scene information for estimating the color of a light source \cite{LuICCV2009}. However, these methods tend to overfit and fail to generalize to all scene types. 

The first attempt to use \acfp{CNN} for solving the illuminant estimation problem was established by  Bianco et al. \cite{22}, where they adopted a \ac{CNN} architecture operating on small local patches to overcome the data shortage. In the testing phase, a map of local estimates is pooled to obtain one global illuminant estimate using median or mean pooling. Hu et al. \cite{44} introduced a pooling layer, namely confidence-weighted pooling. In their fully convolutional network, they incorporate learning the confidence of each patch of the image in an end-to-end learning process. Patches in an image can carry different confidence weights according to their estimated accuracy in predicting the illumination. Shi et al. \cite{DSN} proposed a network with two interacting sub-networks  to  estimate the illumination. One sub-network, called the hypothesis network, is used to generate multiple plausible illuminant estimations depending on the patches in the scene. The second sub-network, called the selection network, is trained to select the best estimate generated by the first sub-network. Inspired by the success of \acfp{GAN} in image to image translation\cite{Isola2017ImagetoImageTW}, Das et al. formulated the illumination estimation task as an image-to-image translation task \cite{Das2018ColorCB} and used a \ac{GAN} to solve it. However, these \ac{CNN}-based methods suffer from certain weaknesses: computational complexity and disconnection with both the illumination assumption\cite{NUS} and the prior static methods, e.g., Grey-World \cite{4} and White-Patch \cite{2}. This paper attempts to cure these drawbacks by proposing a novel \ac{CNN} approach,  \ac{BoCF}, which discards the global spatial information in agreement with \cite{NUS} and \cite{5}, and is competitive with the training-based methods while using only 5\% of the parameters.

\subsection{ Bag-of-Features Pooling } \label{sec:bofp}

Passalis and Tefas proposed a Bag-of-Features Pooling  (BoFP) layer \cite{bof,bofp}, which is a neural extension of the Bag-of-Features model (BoF). BoFPL can be combined with convolutional layers to form a powerful architecture which can be trained end-to-end using the regular back-propagation algorithm \cite{bofp,boftime}. In this work, we use this pooling technique to learn the codebook of color features. Thus, the naming Bag of Color Features (BoCF). This pooling discards all the global spatial information and outputs a fixed length histogram representation. This allows us to reduce the large number of parameters usually needed when linking convolutional layers to fully connected layers. Furthermore, discarding global spatial information forces the network to learn to extract the illumination without global spatial inference, thus improving model robustness and adhering to the illumination assumption \cite{NUS}. 
As an additional novel feature to the prior works using Bag-of- Features Pooling \cite{bofp,boftime}, we propose introducing an attention mechanism to enables the model to discard noise and focus only on relevant parts of the input presentation. To the best of our knowledge, this is the first work which combines attention mechanisms with Bag-of-Features Pooling. 

\subsection{Attention mechanisms} \label{sec:atten}

Attention mechanisms were introduced in \acf{NLP} \cite{FIRst} for sequence-to-sequence (seq2seq) models in order to tackle the problem of short-term memory faced in machine translators. They allow a machine translator to see the full information contained in the original input and then generate the proper translation for the current word. More specifically, they allow the model to focus on local or global features, as needed. Self-attention \cite{attentionyouneed}, also known as intra-attention, is an attention mechanism relating different positions of a single sequence in order to compute a representation of the same sequence. In other words, the attention mask is computed directly from the original sequence. This idea has been exported to many other problems in NLP and computer vision such as machine reading \cite{Cheng2016LongSM}, text summarization \cite{exploration,machine-translation}, and image description generation \cite{Show}. In \cite{Show}, self-attention is applied to an image to enable the network to generate an attention mask and focus on the region of interest in the original image. 

Attention in deep learning can be broadly interpreted as a mask of importance weights. In order to evaluate the importance of a single element, such as a pixel or a feature in general, for the final inference, one can form an attention vector by estimating how strongly the element is correlated with the other elements and use this attention vector as a mask when evaluating the final output \cite{Show}. Let $\textbf{x} = [x_1 , ..., x_n ] \in \mathbb{R}^n$ be a vector. The goal of a self-attention mechanism is to learn to generate a mask vector $\textbf{v} \in \mathbb{R}^n $ depending only on \textbf{x}, which encodes the importance weights of the elements of \textbf{x}. Let $f$ be a mapping function between \textbf{x} and \textbf{v}. The dependency can be expressed as follows: 

\begin{equation}
            \textbf{v} = f(\textbf{x}) = [v_1, ..., v_n],
\end{equation}
under the constraint:

\begin{equation}
            \sum_{i=1}^n v_i = 1,
\end{equation}

After computing the mask vector v, the final output of the attention layer y is computed as follows: 

\begin{equation}
            \textbf{y} = [y_1, ..., y_n] = [x_1v_1, ..., x_nv_n],
\end{equation}
The concept of attention, i.e., focusing on particular regions to extract the illumination information in color constancy, can be rooted back to many statistical approaches. For example, White-Patch reduces this region to the pixel with the highest RGB values. Other methods, such as \cite{8} focus on the bright areas in the captured scene, called specular highlights. Instead of making such a strong assumption on the relevant regions, in \ac{BoCF} we allow the model to learn to extract these regions dynamically. To the best of our knowledge, this is the first work, which uses attention mechanisms in the color constancy problem.

\section{Proposed approach } \label{sec:Proposed}

\begin{figure*}[t]
\centering
\includegraphics[scale=0.08]{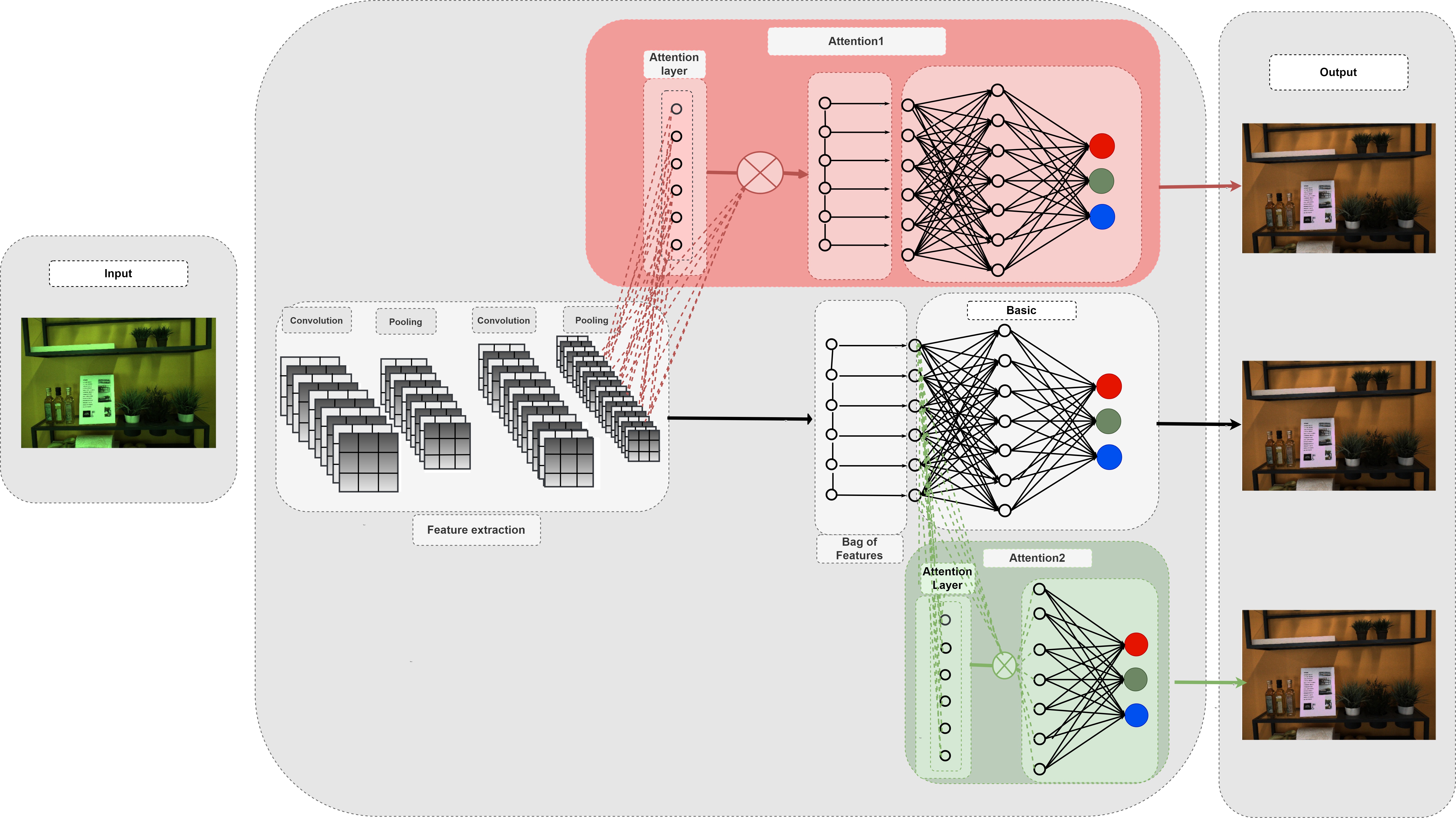}
% where an .eps filename suffix will be assumed under latex, 
% and a .pdf suffix will be assumed for pdflatex; or what has been declared
% via \DeclareGraphicsExtensions.
\caption{Proposed approach (basic, no attention) along with attention variants (Attention1 and Attention2).}
\label{blockmodel}
\end{figure*}

In order to reduce the number of parameters needed to learn the illumination \cite{44,DSN}, we propose a novel color constancy approach based on the Bag-of-Features Pooling \cite{bofp}, called herein the BoCF approach. The proposed approach along with the novel attention variants is illustrated in Figure \ref{blockmodel}. The proposed model has three main blocks, namely the feature extraction, the Bag of Features, and the illumination estimation blocks. In the first block, a nonlinear transformation of a raw image is obtained. In the second block, a histogram representation of this transform is compiled. This histogram is used in the third block to estimate the illumination. 

\subsection{Feature extraction}

The feature extraction algorithm takes a raw image as input and outputs a nonlinear transformation representing the image features. A CNN is used in this block. \acp{CNN} are known for their ability to extract relevant features directly from raw images. Technically, any CNN architecture can be used in this block. However, we observed in our experiments that only two convolutions followed by downsampling layers, e.g., max-pooling yields satisfactory results. This is in accordance with the assumption of statistical methods that the second order information is enough to estimate the illumination \cite{NUS,5}. 

After a raw image is fed to the feature extraction block, the output of the last convolutional layer is used to extract feature vectors that are subsequently fed to the next block. The number of extracted feature vectors depends on the size of the feature map and the used filter size as described in \cite{bofp}. 
\subsection{Bag of Features }
The Bag of Features is essentially the codebook (dictionary) learning component. The output features of the previous block are pooled using the Bag-of-Features Pooling and mapped to a final histogram representation. During training, the network optimizes the codebook using the traditional back-propagation. The output of this block is a histogram of a fixed size, i.e., the size of the codebook, which is a hyper-parameter that needs to be carefully tuned to avoid over-fitting. This approach discards all global spatial information. 
As described in \cite{bofp}, the Bag-of-Features Pooling is composed of two sub-layers: an RBF layer that measures the similarity of the input features to the RBF centers and an accumulation layer that builds the histogram of the quantized feature vectors. The normalized output of each RBF neuron can be expressed as 

\begin{equation}
            [\Phi(\textbf{x})]_k = \frac{exp(- || \textbf{x} - \textbf{v}_k || / \rho_k )}{ \sum_m exp(- || \textbf{x} - \textbf{v}_m || / \rho_m )  }  ,
\end{equation}
where \textbf{x} is a feature vector, $v_k$ is the center of the k-th RBF neuron, exp is the exponential function, and $\rho_k$ is a scaling factor. The output of the RBF neurons is accumulated in the next layer, compiling the final representation of each image: 

\begin{equation}
          \textbf{S} = \frac{1}{ N } \sum_j \Phi(\textbf{x}_j)  ,
\end{equation}
where N is the number of feature vectors extracted from the last convolutional layer for the image. 

\subsection{Illumination Estimation}
The Bag of Features layer receives a transformation of the raw image and compiles its histogram representation. Then, this histogram is fed to a regressor that estimates the illu- mination. In this work, a multi-layer perceptron with one hidden layer is used for this purpose, although any other estimator with differentiable loss function can be used.

Let $\textbf{x}   \in \mathbb{R}^n $ be the histogram compiled by the second block. The intermediate layer output  $\textbf{h}   \in \mathbb{R}^m $ can be computed as follows

\begin{equation}
          \textbf{h} = \varphi(\textbf{W}_1\textbf{x} +\textbf{b}_1)  ,
\end{equation}
where $\textbf{W}_1 \in \mathbb{R}^{n\times m}$ is the weight matrix, $b_1 \in  \mathbb{R}^{m} $ is the bias vector, and $\varphi$ is the Rectiﬁed Linear Units (ReLU) activation function \cite{Agarap2018DeepLU}. The final estimate $\textbf{I} \in \mathbb{R}^{3} $ is computed as follows

\begin{equation}
          \textbf{I} = \phi(\textbf{W}_2\textbf{h} +\textbf{b}_2)  ,
\end{equation}
where $\textbf{W}_2 \in \mathbb{R}^{m\times3}$ is the weight matrix, $b_2 \in  \mathbb{R}^{3} $ is the bias vector, and $\phi$ is the \textit{softmax} activation function defined by 

\begin{equation}
          \phi(a_i) = \frac{exp(a_i)}{ \sum_j a_j}  ,
\end{equation}

\subsection{Attention mechanism for \ac{BoCF}}\label{sec:attenlam}

 We introduce a novel attention mechanism in the \ac{BoCF} model to enable the algorithm to dynamically learn to focus on a specific region of interest in order to yield a confident output. We combine self-attention, described in Section \ref{sec:atten}, with the Bag-of-Features Pooling for the color constancy problem. We propose two variants of this mechanism which can be applied in our model. For the mapping function f in (Eq. 4), we use a fully connected layer with \textit{softmax} activation. 
 
In the first variant, we propose to apply attention on the nonlinear transformation of the image after the feature extraction block. This enables the model to learn to ’attend’ the region of the interest in the mapping and to reduce noise before pooling. By applying attention in this stage, the number of parameters will rise exponentially as we need as many parameters as features. 

In the second variant, we propose to apply the attention mechanism on the histogram representation of the \ac{BoCF}, i.e., after the global spatial information is discarded. This enables the model to dynamically learn to ’tend’ to the relevant parts of the histogram which encode the illuminant information. In this variant, the attention mask size is equal to the size of the histogram. Thus, the number of additional parameters is relatively small. Following the notations of (4) and (5), $\textbf{x}   \in \mathbb{R}^n $   is the histogram representation and $\textbf{v}   \in \mathbb{R}^n $  is the attention mask is obtained via the fully connected layer as follows: 

\begin{equation}
          \textbf{v} = \varphi(\textbf{W}\textbf{x} +\textbf{b})  ,
\end{equation}
where $\textbf{W} \in \mathbb{R}^{N\times N}$ is a weight matrix, $\textbf{b} \in  \mathbb{R}^{N} $ is the bias.

Using \textit{softmax} as $\phi$ ensures that the masking constraint defined in (Eq. 5) is not violated. Finally, \textbf{y}, the final output of the attention mechanism, is computed using the following equation
\begin{equation}
          \textbf{y} = \lambda( \textbf{v} \odot \textbf{x} ) + (1 - \lambda) \textbf{x},
\end{equation}
where $\odot$ is the element wise product operator and $\lambda \in  \mathbb{R}$ is a weighting parameter between the masked histogram $\textbf{v} \odot \textbf{x}  $ and the original histogram $\textbf{x}$. $\lambda$ is a learnable parameter in our model. Not using $\lambda$ and outputting only the masked histogram is also another option. However, we determined experimentally that outputting the weighted sum of both the original and the masked version is more robust and stable for the gradient-based optimizers, since it is less susceptible to random initialization weights of the attention. 

Parameter $\lambda$ can be optimized using the gradient decent in the back-propagation process along with the rest of the parameters. Its gradient with respect to the output of the attention block can be obtained via the following equation 
\begin{equation}
         \frac{\partial \textbf{y}}{\partial \lambda} = \textbf{v} \odot \textbf{x} - \textbf{x},
\end{equation}

\section{Experimental setup}
\label{sec:setup}

In  this  section, we  present the  experimental setup used in this work. In Subsection \ref{subsec:data}, we introduce the datasets used to test our models. In Subsection  \ref{subsec:net}, we report the network architectures of the three blocks used in \ac{BoCF}. In Subsection \ref{subsec:eval}, we detail the evaluation process followed in our experiments. Finally, the evaluation metrics used are briefly described in Subsection \ref{subsec:RAE}.

\subsection{Image datasets}\label{subsec:data}

\subsubsection{ColorChecker RECommended dataset} 
ColorChecker RECommended dataset \cite{Hemrit2018RehabilitatingTC} is a publicly available updated version of Gehler-Shi dataset \cite{47}\footnote{http://www.cs.sfu.ca/~colour/data/shi\_gehler/} with a proposed (recommended) ground truth to use for evaluation. This dataset contains 568 high-quality indoor and outdoor images acquired by two cameras: Canon 1D and Canon 5D.

Similar to the works in \cite{ 22,44,DSN,HoldGeoffroy2017DeepOI}, for Color Cheker REComended dataset, we used three-fold cross validation to evaluate our algorithms.

\subsubsection{NUS-8 Camera Dataset}
NUS-8 is a publicly available dataset\footnote{http://cvil.eecs.yorku.ca/projects/public\_html/illuminant/illuminant.html}, containing 1736 raw images from eight different camera models. Each camera has about 210 images. Following previous works \cite{NUS,44}, we perform tests on each camera separately and report the mean of all the results for each evaluation metric. As a result, although the total number of images in NUS-8 dataset is large, each experiment involves using only 210 images for both training and testing. 

\subsubsection{INTEL-TUT2}

\begin{figure}[t]
\centering
\includegraphics[scale=0.5]{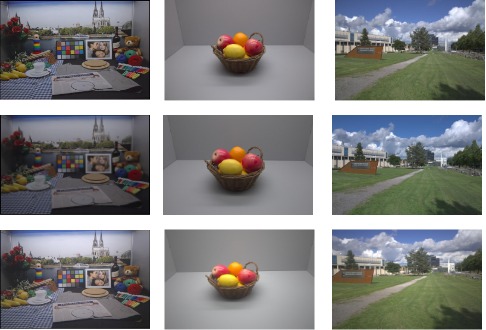}
% where an .eps filename suffix will be assumed under latex, 
% and a .pdf suffix will be assumed for pdflatex; or what has been declared
% via \DeclareGraphicsExtensions.
\caption{Samples from INTEL-TUT2 dataset. The rows contain samples images taken by Canon, Nikon, and mobile cameras, respectively, while the columns contain images from \textit{lab printouts}, \textit{lab real} scenes, and \textit{field}, respectively. }
\label{CNM}
\end{figure}

\begin{figure}[t]
\centering
\includegraphics[scale=0.5]{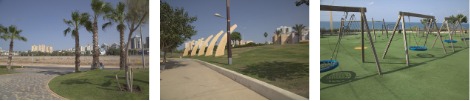}
% where an .eps filename suffix will be assumed under latex, 
% and a .pdf suffix will be assumed for pdflatex; or what has been declared
% via \DeclareGraphicsExtensions.
\caption{Samples from \textit{field2} set specific for Canon in INTEL-TUT2 dataset.}
\label{field2}
\end{figure}

INTEL-TUT2\footnote{http://urn.fi/urn:nbn:fi:csc-kata20170901151004490662} is the second version of the publicly available  INTEL-TUT dataset \cite{17}.  The main particularity of this dataset is that it contains a large number of images taken by several cameras from different scenes. We use this dataset for an extreme testing protocol, the third protocol described in \cite{17}.  The models are trained with images acquired by one camera and containing one type of scene and tested on the other cameras and the other scenes. This extreme test is useful to show the robustness of a given model and its ability to generalize across different cameras and scenes.

INTEL-TUT2 contains images acquired with three different cameras, namely Canon, Nikon, and, Mobile. For each camera, the images are divided into four sets: \textit{field} (144 images per camera), \textit{lab printouts} (300 images per camera), \textit{lab real scenes} (4 images per camera), and \textit{field2}. The last set \textit{field2} concerns only Canon and it has a total of 692 images. Figure \ref{CNM} shows some samples from the  \textit{field}, \textit{lab printouts}, and \textit{lab real scenes} sets of the three cameras, while Figure \ref{field2} displays samples from \textit{field2} related to Canon camera. 

We  used only  Canon \textit{field2} set  for  training  and  validation  (80\%  for training  and  20\% for validation). We constructed two test sets. The first one, called \textit{field} in this work, contains all the field images taken by  the  other  camera  models,  i.e.,  Nikon  and  Mobile.  The second set, called  \textit{non-field} in this work, contains all the non-field  images  acquired  by  Nikon  and  Mobile. Comparing the performance  on these two sets allowed us to test both scene and camera invariance of the model. As we are using different camera models in same experiments, the variation of camera spectral sensitivity needs to be discounted. For this purpose, we use Color Conversion Matrix (CCM) based preprocessing\cite{25} to learn the $3\times3$ \ac{CCM} matrices for each camera pair.

\subsection{Network architectures}  \label{subsec:net}
The \ac{BoCF} network is composed of three blocks: the feature extraction, the Bag of Features, where the Bag-of-Features Pooling is applied, and the illumination estimation blocks as described in Section \ref{sec:Proposed}. The feature extraction block consists of convolution layers followed by max pooling operators. We experiment with different number of layers two and three. Thirty convolution filters of size $4 \times 4$ are used in both layers. Max-pooling with a window size 2 is applied in both layers. For the codebook size, i.e., number of RBF neurons in the Bag of Features block, we experiment with 3 different values 50, 150 and 200. The illumination estimation block consists of 2 fully connected layers, the first (hidden layer) has a size of  40 and it takes as input the histogram representation and the second one (final output) has of size 3 to output the illumination.

\subsection{Evaluation procedure} \label{subsec:eval}
To evaluate the proposed approach, we used 2 sets of experiments. In the first set, we evaluate different variants of the model to study the effect  of the hyper-parameters and validate the effectiveness of each component in our model by conducting ablation studies.  For this purpose, we used ColorChecker RECommended dataset. In the second set of experiments, we compared our approach with current state-of-the-art approaches on the three datasets.

For all testing scenarios, we augmented the datasets using the following process: As the size of the original raw images is high, we first randomly cropped $512\times512$ patches of each image. This ensured getting meaningful patches. The crops were then rotated by a random angle between -30\degree and +30\degree. Finally, we rescaled the RGB values of each patch and its corresponding ground truths by random factor in the range of [0.8, 1.2]. Before feeding the sample to the network, we down-sampled it to $227\times227$. In testing, the images are resized to $227\times227$ to fit the network model.

Our network was implemented in Keras \cite{chollet2015keras} with Tensorflow backend \cite{45381}. We trained our network end-to-end by back-propagation. For optimization, Adam \cite{Kingma2015AdamAM} was employed with a batch size of 15 and a learning rate of $3 \times 10^{−4}$. The model was trained on image patches of size $227\times227$ for 3000 epochs. The centers of the dictionary were initialized using the k-means algorithm as described in \cite{bofp}. The parameter $\lambda$, discussed in Section \ref{sec:attenlam}, was initialized as 0.5.

 \subsection{Evaluation metrics} \label{subsec:RAE}
We report the mean of the top 25\%, the mean, the median, Tukey's trimean, and the mean of the worst 25\% of the \acf{RAE} \cite{21} between the ground truth illuminant and the estimated illuminant, defined as
\begin{equation}
     \text{\ac{RAE}}(\rho^{gt},\rho^{Est})= \cos^{-1} ({ \frac{ \rho^{gt} \rho^{Est}}{\| \rho^{gt} \| \|\rho^{Est} \| } }), 
\end{equation}
where $\rho^{gt}$ is the ground truth illumination for a given image and $\rho^{Est}$ is the estimated illumination.% The mean of the worst 25\% is important as it reflects more about the generalization ability of the model. 

\section{Experimental results} \label{sec:res}
In this section, we provide the experimental evaluation of the proposed method and its variants. In Subsection \ref{subsec:resint}, different topologies for the three blocks of \ac{BoCF} are evaluated on the ColorChecker RECommended dataset and the effect of each block in our model is examined by reporting the results of the ablation studies. In Subsection  \ref{subsec:RecResults}, we compare the performance of the proposed models with different state-of-the-art algorithms over the three datasets.

\subsection{ \ac{BoCF} performance evaluation} \label{subsec:resint}

% Example table
\begin{table*}[h]
 \centering	
	\caption{Comparison of different variants of the proposed \ac{BoCF} approach on ColorChecker RECommended dataset}
	\label{tab:tableinternal}
	\begin{tabular}{l|c|ccccc} % four columns, alignment for each
		\hline
Method & \# par  & Best 25\% & Mean & Med. & Tri. & Worst 25\%  \\	
		\hline
\ac{BoCF}(2conv+50 words + no attention) & 13k& 0.4 & 2.2 & 1.6 & 2.0  & 5.3 \\
\ac{BoCF}(2conv+150 words + no attention) & 20k& 0.3 & 2.1 & 1.5 & 1.6  & 5.1 \\
\ac{BoCF}(2conv+200 words + no attention) & 23k& 0.3 & 2.0 & 1.5 & 1.6  & 5.2 \\
\ac{BoCF}(3conv+150 words+ no attention) & 37k & 0.3 & 2.2 & 1.4 & 1.8  & 5.1 \\
% \ac{BoCF}(SqueezeNet \cite{squueze}+150 words+ no attention) & 37k & 0.4 & 2.2 & 1.5 & 1.7  & 5.2 \\

 		\hline

\ac{BoCF}(2conv+50 words + attention1) & 369k & 0.4 & 2.0 & 1.3 & 2.0  & 5.1 \\
\ac{BoCF}(2conv+150 words + attention1) & 376k & 0.3 & 2.0 & 1.3 & 1.5  & 4.7 \\
\ac{BoCF}(2conv+200 words + attention1) & 380k & 0.3 & 2.0 & 1.2 & 1.5  & 5.0 \\
		\hline
\ac{BoCF}(2conv+50 words + attention2) & 15k& 0.4 & 2.2 & 1.5 & 1.6  & 5.1 \\
\ac{BoCF}(2conv+150 words + attention2) & 43k& 0.3 & 2.0 & 1.2 & 1.4  & 4.8 \\
\ac{BoCF}(2conv+200 words + attention2) & 63k& 0.3 & 2.0 & 1.3 & 1.5  & 4.8 \\

	\end{tabular}
\end{table*} 
We first evaluated the  accuracy of the different variants of \ac{BoCF} on ColorChecker RECommended dataset. Table \ref{tab:tableinternal} presents the comparative results for \ac{BoCF} using different topologies in the three blocks. We evaluate the model using different dictionary sizes in the second block (codewords), different numbers of convolution layers in the first block, and with/without attention.

Table \ref{tab:tableinternal} shows that the dictionary size in the Bag-of-Features Pooling block significantly affects the overall performance  of the model. Using a larger codebook results in  higher risk of overfitting to the training data, while using a smaller codebook size restricts the model to only few codebook centers which can decrease the overall performance  of the model. Thus, the choice of this hyperparameter is critical for our model. The findings in Table \ref{tab:tableinternal} confirm this effect and highlights the importance of this hyperparameter. By comparing the model performance using different dictionary sizes, we can see that a dictionary of size 150 yields the best compromise between the number of parameters and the overall performance.

Using three convolutional layers instead of two in the first block yields slightly better median errors and worse trimean errors. However, to keep the model as shallow as possible, we opt for the two convolution layers. 

Table \ref{tab:tableinternal} shows that models equipped with an attention mechanism perform better than models without attention almost consistently  across all error metrics. This is expected as attention mechanisms allow the model to focus on relevant parts only and as a result, the model becomes more robust to noise and to inadequate features. The performance  boost obtained by both attention variants is more highlighted in terms of the median and trimean errors compared to the non-attention variant.

By comparing the performance achieved by the two attention variants, we note that the first attention variant yields in a better performance in terms of worse 25\% error rate, while the second variant yields a better median and trimean error rates. It should also be remembered that the first variant applies attention over the feature map output of the first convolutional block. Thus, it dramatically increases the number of model parameters (over 20 times) compared to the second variant (doubling the number of parameters) which applies the attention over the histogram. 

\begin{figure*}[h]
\centering
\includegraphics[scale=1]{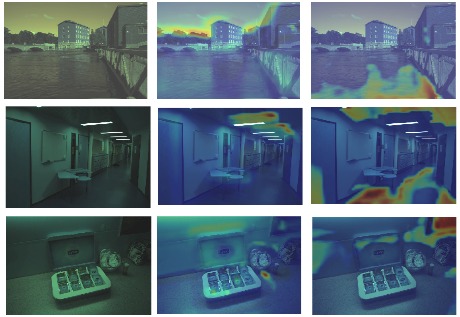}
% where an .eps filename suffix will be assumed under latex, 
% and a .pdf suffix will be assumed for pdflatex; or what has been declared
% via \DeclareGraphicsExtensions.
\caption{Attention mask visualization \cite{raghakotkerasvis} for three samples from INTEL-TUT2 dataset. The first column contains the input image. The second one illustrates the attention mask generated by the first attention variant overlaid on the input image. The last column contains the attention masks generated by the second variant of the attention overlaid on the input image. Gamma correction was applied for visualization.}
\label{att}
\end{figure*}

Figure \ref{att} presents a  visualization of the attention weights\cite{raghakotkerasvis} for both attention variants. The heat maps demonstrate which regions of the image each model pays attention to so as to output a certain illumination. We note a large difference between both attentions. The first attention variant tends to focus on regions with dense edges and  sharp shapes, while the second model focuses on uniform regions to estimate the illumination.

\subsection*{Ablation studies} \label{subsec:abla}
To examine the effect of each block in our proposed approach, we conduct ablation studies on the colorChecker RECommended dataset. Table \ref{tab:ablation} reports the results of the basic \ac{BoCF} approach, the results achieved by removing the feature extraction block, and the results obtained by removing the estimation block, i.e., replacing the fully connected layer in the estimation block with a simple regression, 
%and the result obtained by replacing the BoF block by a global average pooling layer \cite{Lin2014NetworkIN}.
We note that removing any block significantly decreases the overall performance of our models. 

Comparing the model with and without the feature extraction block, we note a large drop in performance especially in terms of the worst 25\% error rates, i.e., 1.8\degree drop compared to 0.6\degree drop  when  the estimation block is removed.
%We also note that BoF pooling performs better than  global average pooling across all metrics except the median.

\begin{table}[h]

%if using array.sty, it might be a good idea to tweak the value of
%\extrarowheight as needed to properly center the text within the cells
	\caption{Results of the ablation studies for the \ac{BoCF}  over the RECommended Color Checker Dataset. \ac{BoCF}  is the basic \ac{BoCF}  composed of the three blocks. In \ac{BoCF} -1, the feature extraction block is removed, while in \ac{BoCF} -2 the fully connected layer in the estimation block is substituted with a linear regression.% In \ac{BoCF} -3, the BoF pooling layer is replaced with a global average pooling layer.
	}
	\label{tab:ablation}
\centering
% Some packages, such as MDW tools, offer better commands for making tables
% than the plain LaTeX2e tabular which is used here.
\begin{tabular}{l|ccccc} % four columns, alignment for each
		\hline
Method   & Best 25\% & Mean & Med. & Tri. & Worst 25\%  \\	

		\hline
\ac{BoCF} &  0.3 & 2.1 & 1.5 & 1.6  & 5.1 \\
\ac{BoCF}-1 &  0.4 & 2.9 & 1.9 & 2.2  & 6.9 \\
\ac{BoCF}-2 &  0.5 & 2.4 & 1.7 & 1.7  & 5.7 \\
%\ac{BoCF}-3 & 0.4 &  2.2 &  1.5&  1.7 &  5.2 \\

	\end{tabular}
\end{table}

\subsection{Comparisons against state-of-the-art} \label{subsec:RecResults}
We compare our \ac{BoCF} approach with the state-of-the-art methods on ColorChecker RECommended, NUS-8, and INTEL-TUT2 datasets, which have been widely adopted as benchmark datasets in the literature. 
Tables \ref{tab:rec2}, \ref{tab:nus}, and \ref{tab:intel} provide quantitative results for ColorChecker RECommended, NUS-8, and INTEL-TUT2 datasets, respectively. We provide results for the static methods Grey-World, White-Patch, Shades-of-Grey, and General Grey-World. The parameter values $n$, $p$, $\rho$ are set as described in \cite{5}. In addition, we compare against Pixel-based Gamut, Bright Pixels, Spatial Correlations, Bayesian Color Constancy \cite{47}, and six convolutional approaches: Deep Specialized Network for Illuminant Estimation (DS-Net) \cite{DSN}, Bianco \ac{CNN}  \cite{22},  Fast Fourier Color Constancy \cite{46440}, Convolutional Color Constancy\cite{Barron2015ConvolutionalCC}, Fully Convolutional Color Constancy With Confidence-Weighted Pooling (FC4) \cite{44}, and Color Constancy GANs (CC-GANs) \cite{Das2018ColorCB}. The results for ColorChecker RECommended and NUS-8 datasets were taken from related papers \cite{Das2018ColorCB,44}. 

From Recommended Color Checker and NUS-8 datasets results in Tables \ref{tab:rec2} and \ref{tab:nus}, we note that learning-based methods usually outperform statistical-based methods across all error metrics. This can be explained by the fact that statistical approaches rely on some assumptions in their model. These assumptions can be violated in some testing samples  which results in high error rates especially in terms of the worst 25\% errors. 

Table \ref{tab:rec2} shows that the proposed method \ac{BoCF} and its variants achieve competitive results on Recommended Color Checker dataset. The only models performing slightly better than \ac{BoCF} are  FC4(SqueezeNet) and DS-Net. By comparing the number of parameters required by each model given in Table  \ref{tab:parameters}, we see that \ac{BoCF} achieves very competitive results, while using less than 1\% of the parameters of FC4(SqueezeNet) and less than 0.1 \% of the parameters of DS-Net.

Compared to Bianco's \ac{CNN}, we note that our model performs better across all error metrics except for the worst 25\% error metric. Bianco \ac{CNN} operates on patches instead of the full image directly and this makes it more robust but, at the same time, it increases its time complexity as the network has to estimate many local estimates before outputting the global one. 
% Example table
\begin{table}[h]
\renewcommand{\arraystretch}{1}
 \centering	
 %\footnotesize\setlength{\tabcolsep}{2pt}
	\caption{Number of parameters of different \ac{CNN}-based approaches}
	\label{tab:parameters}
	\begin{tabular}{l|r} % four columns, alignment for each
		\hline
Method         &  \# parameters  \\	
\hline
Bianco  \cite{22} &   154k   \\
Fc4(SqueezeNet)   \cite{44} &   1.9M  \\
FC4 (AlexNet)   \cite{44}& 3.8M \\
DS-Net \cite{DSN} &   17.3M   \\
\hline
\ac{BoCF}(2conv+150 words + no attention) &  20k \\
\ac{BoCF}(2conv+150 words + attention1) &  376k \\
\ac{BoCF}(2conv+150 words + attention2) &  43k \\

	\end{tabular}
\end{table}

Results for NUS-8 dataset are similar to their counter parts on ColorChecker RECommended, as illustrated in Table  \ref{tab:nus}. Our models achieve comparable results with  FC4 and overall better results compared to DS-Net across all error metrics. Bianco's \ac{CNN} outperforms all the other \ac{CNN}-based methods. As discussed earlier, this can likely be explained by the fact that Bianco operates on patches while \ac{BoCF} and FC4 produce global estimates directly.   

Table \ref{tab:intel} reports the comparative results achieved on INTEL-TUT2 dataset. We note that all the error rates are high as this is an extreme testing senario. The models are trained and validated using only one type of scene (\textit{field2} set) acquired by one camera model (Canon) and then evaluated over different scene types and different camera models not seen during the training as described in Section \ref{subsec:eval}. The proposed \ac{BoCF} model achieves better overall performance  compared to Bianco's \ac{CNN} and \ac{C3AE} methods and competitive results compared to FC4. 

By comparing the performance  achieved by \ac{BoCF} with and without attention, we note both the attention mechanisms proposed in this paper significantly boost the performance of our model for all datasets. It should also be mentioned that despite requiring much less parameters, the second variant of our attention model, where the attention is applied over the histogram representation, performs slightly better than the first variant, where the attention is applied over the feature extraction block.

%\textbf{
%The visualization of the attention weights clearly demonstrates which regions of the image the model %pays attention to so as to output a certain word.
%This result ties well with previous studies wherein..}

\begin{table*}[t]
% increase table row spacing, adjust to taste

\renewcommand{\arraystretch}{1}

%if using array.sty, it might be a good idea to tweak the value of
%\extrarowheight as needed to properly center the text within the cells
	\caption{Results of \ac{BoCF} approach and comparative methods  on the RECommended Color Checker Dataset. }
	\label{tab:rec2}
\centering

% Some packages, such as MDW tools, offer better commands for making tables
% than the plain LaTeX2e tabular which is used here.
\begin{tabular}{l|c|cccccc} % four columns, alignment for each

\multirow{2}{*}{Method}                                    & \multicolumn{2}{c}{Type}                                        & \multicolumn{1}{l}{\multirow{2}{*}{Best 25\%}} & \multicolumn{1}{l}{\multirow{2}{*}{Mean}} & \multicolumn{1}{l}{\multirow{2}{*}{Med.}} & \multicolumn{1}{l}{\multirow{2}{*}{Tri.}} & \multicolumn{1}{l}{\multirow{2}{*}{Worst 25\%}} \\
                                                           & \multicolumn{1}{c|}{statistic-based} & learning-based            & \multicolumn{1}{l}{}                           & \multicolumn{1}{l}{}                      & \multicolumn{1}{l}{}                      & \multicolumn{1}{l}{}                      & \multicolumn{1}{l}{}  
                                                           \\ \hline
                                                           
Grey-World \cite{4}  &\checkmark & --  &5.0 & 9.7 & 10 & 10 & 13.7 \\
White-Patch \cite{2}  &\checkmark&  -- & 2.2 & 9.1 & 6.7 & 7.8 & 18.9 \\

Shades-of-Gray \cite{shades}  &\checkmark &  --& 2.3 & 7.3 & 6.8 & 6.9 & 12.8 \\

General-gray world \cite{4} &\checkmark&  -- & 2.0 & 6.6 & 5.9 & 6.1 & 12.4 \\

Pixel-based Gamut \cite{644} & \checkmark& -- & 1.7 & 6.0 & 4.4 &4.9 &12.9 \\ 
Top-down \cite{high}  & \checkmark& -- & 2.3 & 6.0 & 4.6 & 5.0 & 10.2 \\ 
Spatial Correlations  \cite{chzcc2011} &\checkmark&  --& 1.9 & 5.7 & 4.8 & 5.1 & 10.9 \\ 

Bottom-up \cite{high} & \checkmark & -- & 2.3 & 5.6 & 4.9 & 5.1 & 10.2 \\ 

Edge-based Gamut \cite{644} & \checkmark& -- &  0.7 & 5.5 & 3.3 & 3.9 & 13.8 \\ 

		\hline

 CC-GANs (Pix2Pix) \cite{Das2018ColorCB} & --&\checkmark & 1.2 & 3.6 & 2.8& 3.1& 7.2 \\ 
 CC-GANs (CycleGAN) \cite{Das2018ColorCB} & --&\checkmark & 0.7 & 3.4 & 2.6& 2.8& 7.3 \\ 
 CC-GANs (StarGAN) \cite{Das2018ColorCB} & --&\checkmark & 1.7 & 5.7 & 4.9&5.2& 10.5 \\

FFCC (model Q) \cite{46440} & --&\checkmark &  0.3 &  2.0 &  1.1&  1.4& 5.1 \\ 
Cheng et al. 2015 \cite{7298702} &  --&\checkmark  & 0.4 & 2.4 &1.7 &1.7 &5.9 \\
DS-Net \cite{DSN} & -- &\checkmark &  0.3 & 1.9 & 1.1 & 1.4 &4.8 \\
CCC\cite{Barron2015ConvolutionalCC} &  --&\checkmark  & 0.3 &  2.0 & 1.2 & 1.4 &4.8 \\
% pilot3 w.oconf.  &112k&  0.51 & 1.98 &1.38& 1.52 & 4.52 \\
% pilot3 w conf.  &189k & 0.44 & 1.85 &1.31 &1.37&4.14 \\
Bianco CNN \cite{22}& -- &\checkmark & 0.8 &2.6 &2.0 &2.1 & 4.0 \\
FC4(SqueezeNet)  \cite{44}& -- &  \checkmark  & 0.4 &  1.7 & 1.2 &  1.3 &   3.8\\

		\hline
\ac{BoCF}(2conv+150 words + no attention) &  -- &\checkmark &  0.3 & 2.1 & 1.5 & 1.6  & 5.1 \\
\ac{BoCF}(2conv+150 words + attention1) &  -- &\checkmark &  0.3 &  2.0 &  1.3 &  1.5  &  4.7 \\
\ac{BoCF}(2conv+150 words + attention2) &  -- &\checkmark &  0.3 &  2.0 &  1.2 &  1.4  & 4.8 \\

	\end{tabular}
\end{table*}

% Note that the IEEE does not put floats in the very first column
% - or typically anywhere on the first page for that matter. Also,
% in-text middle ("here") positioning is typically not used, but it
% is allowed and encouraged for Computer Society conferences (but
% not Computer Society journals). Most IEEE journals/conferences use
% top floats exclusively. 
% Note that, LaTeX2e, unlike IEEE journals/conferences, places
% footnotes above bottom floats. This can be corrected via the
% \fnbelowfloat command of the stfloats package.

\begin{table*}[t]
% increase table row spacing, adjust to taste
\renewcommand{\arraystretch}{1}
%if using array.sty, it might be a good idea to tweak the value of
%\extrarowheight as needed to properly center the text within the cells
	\caption{Results of \ac{BoCF} approach and benchmark methods  on the NUS-8 Dataset.}
	\label{tab:nus}
\centering
% Some packages, such as MDW tools, offer better commands for making tables
% than the plain LaTeX2e tabular which is used here.
\begin{tabular}{l|c|cccccc} % four columns, alignment for each
\multirow{2}{*}{Method}                                    & \multicolumn{2}{c}{Type}                                        & \multicolumn{1}{l}{\multirow{2}{*}{Best 25\%}} & \multicolumn{1}{l}{\multirow{2}{*}{Mean}} & \multicolumn{1}{l}{\multirow{2}{*}{Med.}} & \multicolumn{1}{l}{\multirow{2}{*}{Tri.}} & \multicolumn{1}{l}{\multirow{2}{*}{Worst 25\%}} \\
                                                           & \multicolumn{1}{c|}{statistic-based} & learning-based            & \multicolumn{1}{l}{}                           & \multicolumn{1}{l}{}                      & \multicolumn{1}{l}{}                      & \multicolumn{1}{l}{}                      & \multicolumn{1}{l}{}  
                                                           \\ \hline
                                                           
Grey-World \cite{4} &\checkmark & --  &0.9 & 4.1 & 3.2 & 3.4 & 9.0 \\
White-Patch \cite{2} &\checkmark&  -- & 1.9 &10.6  & 10.6 & 10.5 & 19.4 \\
Shades-of-Gray \cite{shades}  &\checkmark &  --& 0.8 & 3.4 & 2.6 & 2.7 & 7.4 \\
General-gray world \cite{4} &\checkmark&  -- & 0.7 & 3.2 & 2.4 & 2.5 & 7.1 \\
Pixel-based Gamut \cite{644} & \checkmark& -- & 2.5 & 7.7 & 6.7 &6.9 &14.0 \\ 
Bright Pixels \cite{Joze2012TheRO}  &\checkmark&  --& 0.7 & 3.2 & 2.4 & 2.6 & 7.0 \\ 
Edge-based Gamut \cite{644} & \checkmark& -- &  2.4 & 8.4 & 7.0 & 7.4 & 16.1 \\ 
		\hline
Bayesian \cite{47} &--&  \checkmark &  0.8 & 3.7 & 2.7 & 2.9 & 8.2 \\ 

%FFCC (model Q) & --&\checkmark & 0.3 & 2.0 & 1.1& 1.4& 5.1 \\ 
Cheng et al. 2015 \cite{7298702} &  --&\checkmark  &0.6 & 2.9 &2.0 &2.2 &6.6 \\
DS-Net \cite{DSN} & -- &\checkmark & 0.5 &2.2 &1.5 &1.7 &6.1 \\
CCC\cite{Barron2015ConvolutionalCC} &  --&\checkmark  &0.5 & 2.4 &1.5 &1.7 &5.9 \\
Regression Tree \cite{7298702} &  --&\checkmark  &0.5 & 2.4 &1.6 &1.7 &5.5 \\

% pilot3 w.oconf.  &112k&  0.51 & 1.98 &1.38& 1.52 & 4.52 \\
% pilot3 w conf.  &189k & 0.44 & 1.85 &1.31 &1.37&4.14 \\
Bianco\cite{22} & -- &\checkmark & 0.3 &2.6 &2.0 &2.1 &3.9 \\
FC4(SqueezeNet) \cite{44} & -- &  \checkmark  &0.5 & 2.2 & 1.5 & 1.7 &  5.2\\
FC4(AlexNet) \cite{44} & -- &  \checkmark  &0.5 & 2.1 & 1.6 & 1.7 &  4.8\\

		\hline
\ac{BoCF}(2conv+150 words + no attention) &  -- &\checkmark &  0.6 & 2.5 & 1.6 & 1.8  & 5.6 \\
\ac{BoCF}(2conv+150 words + attention1) &  -- &\checkmark &  0.5 & 2.3 & 1.4 & 1.7  & 5.2 \\
\ac{BoCF}(2conv+150 words + attention2) &  -- &\checkmark &  0.5 & 2.3 & 1.5 & 1.7  & 5.1 \\

	\end{tabular}
\end{table*}

% Example table
\begin{table}[h]
\footnotesize\setlength{\tabcolsep}{4pt}
\renewcommand{\arraystretch}{1}
 \centering	
 %\footnotesize\setlength{\tabcolsep}{2pt}
	\caption{Results of \ac{BoCF} approach and benchmark methods on INTEL-TUT2.}
	\label{tab:intel}
	\begin{tabular}{ll|ccccr} % four columns, alignment for each
		\hline
Method  &  set         & Best \newline 25\% & Mean & Med. & Tri. & W. \newline 25\%  \\		\hline
\hline
Bianco  \cite{22}  &    field   & 1.1 &   4.5 &      3.7 &      3.8 &  9.2  \\
      &    non-field &    1.8 &    6.2 &   5.3 & 5.5 &     12.4  \\ 

      \hline

C3AE \cite{mine} &    field   &        1.6 &        4.4 &         4.0 &    4.2 &          7.9  \\
fine-tuned       &   non-field     &        1.6 &        5.2 &       4.6 &       4.7 &        10.1 \\ 
   \hline 
C3AE  \cite{mine}  &    field   & 2.0 & 6.1 & 5.3 & 5.4 &    10.7\\
composite-loss   &    non-field     & 1.9 &    6.2 & 5.3 &    5.4 &  14.4  \\ 
\hline
FC4   \cite{44}    &   field   &       1.7 &      4.3 &       4.1 &         4.2 &       7.4  \\
      &    non-field     &        1.5 &        4.8 &      4.2 &      4.3 &        9.0  \\ 	
     \hline
     
\ac{BoCF} (150 w)     &    field   &       1.7 &   4.6 &       4.1 &         4.2 &       8.1  \\
No attention     &    non-field     &        1.5 &        4.9 &      4.2 &      4.4 &        9.5 \\ 	
     \hline
     
\ac{BoCF} (150 w)     &    field   &       1.9 &   4.5 &       4.1 &         4.2 &       7.3  \\
attention1     &    non-field     &        1.5 &        4.9 &      4.2 &      4.3 &        9.0 \\ 	
     \hline     
\ac{BoCF} (150 w)     &    field   &       1.7 &   4.4 &       4.1 &         4.2 &       7.5  \\ attention2     &    non-field     &        1.5 &        4.9 &      4.3 &      4.4 &        9.1 \\ 	
	\end{tabular}
\vspace{-4mm}
\end{table}

\section{Discussion}\label{subsec:des}

When comparing our approach to the competing methods, it must be pointed out that our approach can be linked to many previous static-based approaches. In Grey-World\cite{4}, one takes the average of the RGB channels of the image. In the proposed method, this corresponds to using the identity as a feature extractor and using equal weights in the estimation block. This way all the histogram bins will contribute equally in the estimation. White-Patch\cite{2} takes the max across the color channels, which corresponds to giving a high weight to the histogram bin with the highest intensity and giving zero weights to the rest. Grey-edge and its variants\cite{5} correspond to using the first and second order derivatives as a feature extractor. Thus, \ac{BoCF} approach can be interpreted as a learning-based generalization of these statistical based approaches. Instead of using the image directly, we allow the model to learn a suitable non-linear transformation of the original image, through the feature extraction block, and instead of imposing a prior assumption on the contribution of each feature in the estimation, we allow the model to learn the mapping dynamically using the training data.

It is interesting to note that the  attention variants in our approach can be tightly linked to the confidence maps in FC4 \cite{44}. In FC4, confidence scores are assigned to each patch of the image and a final estimate is generated by a weighted sum of the scores and their corresponding local estimates. This way the network learns to select which features contribute to the estimation and which parts should be discard. Similarly, attention mechanism learn to dynamically pay attention to the parts encoding the illumination information and discarding the rest. 

\section{Conclusion}\label{sec:conc}

In this paper, we proposed a novel color constancy method called BoCF, which is composed of three blocks. In first block, called feature extraction, we employ convolutional layers to extract relevant features from the input image. In the second block, we apply Bag-of-Features Pooling to learn a codebook and output of histogram. The latter is fed into the last block, the estimation block, where the final illumination is estimated. This end-to-end model is evaluated and compared with prior works over three datasets: ColorChecker RECommended, NUS-8, and INTEL-TUT2. BoCF was able to achieve competitive results compared to state-of-the-art methods while reducing the number of parameters up to 95\%. In this paper, we also discussed links between the proposed method and statistic based methods and we showed how the proposed approach can be interpreted as a supervised extension of these approaches and can act as a generic framework for expressing existing approaches as well as developing new powerful methods. 

In addition, we proposed combining the Bag-of-Features Pooling with two novel attention mechanisms. In the first variant, we apply attention over the nonlinear transform of the image after the feature extraction block. In the second extension, we apply attention over the histogram representation of the Bag-of-Features Pooling. These extensions are shown to improve the overall performance of our model.

In future work, extensions of the proposed approach could include exploring regularization techniques to ensure diversity in the learned dictionary and improve the generalization capability of the model.

% if have a single appendix:
%\appendix[Proof of the Zonklar Equations]
% or
%\appendix  % for no appendix heading
% do not use \section anymore after \appendix, only \section*
% is possibly needed

% use appendices with more than one appendix
% then use \section to start each appendix
% you must declare a \section before using any
% \subsection or using \label (\appendices by itself
% starts a section numbered zero.)
%

%\appendices
%\section{Proof of the First Zonklar Equation}
%Appendix one text goes here.

% you can choose not to have a title for an appendix
% if you want by leaving the argument blank
%\section{}
%Appendix two text goes here.

% use section* for acknowledgment
\ifCLASSOPTIONcompsoc
  % The Computer Society usually uses the plural form
  \section*{Acknowledgments}
\else
  % regular IEEE prefers the singular form
  \section*{Acknowledgment}
\fi

This  work  was  supported  by  the  NSF-Business Finland  Center  for
Visual and Decision Informatics project (CVDI) project AMALIA, Dno 3333/31/2018, sponsored by
Intel Finland.

% Can use something like this to put references on a page
% by themselves when using endfloat and the captionsoff option.
\ifCLASSOPTIONcaptionsoff
  \newpage
\fi

% trigger a \newpage just before the given reference
% number - used to balance the columns on the last page
% adjust value as needed - may need to be readjusted if
% the document is modified later
%\IEEEtriggeratref{8}
% The "triggered" command can be changed if desired:
%\IEEEtriggercmd{\enlargethispage{-5in}}

% references section

% can use a bibliography generated by BibTeX as a .bbl file
% BibTeX documentation can be easily obtained at:
% http://mirror.ctan.org/biblio/bibtex/contrib/doc/
% The IEEEtran BibTeX style support page is at:
% http://www.michaelshell.org/tex/ieeetran/bibtex/
%\bibliographystyle{IEEEtran}
% argument is your BibTeX string definitions and bibliography database(s)
%\bibliography{IEEEabrv,../bib/paper}
%
% <OR> manually copy in the resultant .bbl file
% set second argument of \begin to the number of references
% (used to reserve space for the reference number labels box)

\bibliographystyle{IEEEtran}

\bibliography{strings}

%\begin{thebibliography}{1}

%\bibitem{IEEEhowto:kopka}
%H.~Kopka and P.~W. Daly, \emph{A Guide to \LaTeX}, 3rd~ed.\hskip 1em plus
 % 0.5em minus 0.4em\relax Harlow, England: Addison-Wesley, 1999.

%\end{thebibliography}

% biography section
% 
% If you have an EPS/PDF photo (graphicx package needed) extra braces are
% needed around the contents of the optional argument to biography to prevent
% the LaTeX parser from getting confused when it sees the complicated
% \includegraphics command within an optional argument. (You could create
% your own custom macro containing the \includegraphics command to make things
% simpler here.)
%\begin{IEEEbiography}[{\includegraphics[width=1in,height=1.25in,clip,keepaspectratio]{mshell}}]{Michael Shell}
% or if you just want to reserve a space for a photo:

 \vskip -1\baselineskip plus -1fil

\begin{IEEEbiographynophoto}{Firas Laakom}
is a doctoral student at Tampere university,Finland. He received his engineering degree from Tunisia Polytechnic School (TPS) in 2018.  His research interests include deep learning, computer vision
and computational intelligence.
\end{IEEEbiographynophoto}
 \vskip -1\baselineskip plus -1fil
\begin{IEEEbiographynophoto}{Nikolaos Passalis}
is a postdoctoral researcher at Tampere University, Finland. He has (co-)authored more than 45 journal and conference papers and contributed one chapter to one edited book. His research interests include machine learning, information retrieval and computational intelligence.
\end{IEEEbiographynophoto}
 \vskip -1\baselineskip plus -1fil
\begin{IEEEbiographynophoto}{Jenni Raitoharju} 
is a postdoctoral research fellow in Unit of Computing Sciences, Tampere University, Finland. She received her PhD in Information Technology in Tampere University of Technology in 2017. Her current projects deal with machine learning and pattern recognition in applications such as bio-monitoring, intelligent buildings, and autonomous boats. She has (co-)authored 11 journal papers and 20 conference papers.\end{IEEEbiographynophoto}
 \vskip -1\baselineskip plus -1fil
\begin{IEEEbiographynophoto}{Jarno Nikkanen}
received his M.Sc. and Dr.Sc.Tech. degrees from Tampere University of Technology in 2001 and 2013, respectively, with subjects in Signal Processing and Software Systems. Jarno has 18 years of industry experience in digital imaging topics, starting at Nokia Corporation in 2000 where he developed and productized many digital camera algorithms, and moving to Intel Corporation in 2011 where he is currently working as Intel Principal Engineer and Imaging Technology Architect. Jarno holds international patents for over 20 digital camera related inventions.\end{IEEEbiographynophoto}
 \vskip -1\baselineskip plus -1fil
\begin{IEEEbiographynophoto}{Anastasios Tefas}
an Associate Professor at the Department of Informatics, Aristotle University of Thessaloniki. He has co-authored 100 journal papers, 215 papers in international conferences and contributed 8 chapters to edited books in his area of expertise. Over 4900 citations have been recorded to his publications and his H-index is 36 according to Google scholar. His current research interests include computational intelligence, deep learning, digital signal and image analysis and retrieval and computer vision.
\end{IEEEbiographynophoto}
 \vskip -1\baselineskip plus -1fil
\begin{IEEEbiographynophoto}{Alexandros Iosifidis}
is an Associate Professor at Aarhus University, Denmark. He has contributed in more than ten R\&D projects financed by EU, Greek, Finnish, and Danish funding agencies and companies. He has co-authored 53 articles in international journals and 78 papers in international conferences proposing novel Machine Learning techniques and their application in a variety of problems. Dr. Iosifidis is a Senior Member of IEEE and he served as an Officer of the Finnish IEEE Signal Processing-Circuits and Systems Chapter.\end{IEEEbiographynophoto}

 \vskip -1\baselineskip plus -1fil
\begin{IEEEbiographynophoto}{Moncef Gabbouj}
received his MS and PhD degrees in EE from Purdue University, in 1986 and 1989, respectively. Dr. Gabbouj is Professor of Signal Processing at the Department of Computing Sciences, Tampere University, Finland. His research interests include Big Data analytics, multimedia analysis, artificial intelligence, machine learning, pattern recognition, nonlinear signal processing, video processing and coding. Dr. Gabbouj is a Fellow of the IEEE and member of the Academia Europaea and the Finnish Academy of Science and Letters.\end{IEEEbiographynophoto}

% You can push biographies down or up by placing
% a \vfill before or after them. The appropriate
% use of \vfill depends on what kind of text is
% on the last page and whether or not the columns
% are being equalized.

%\vfill

% Can be used to pull up biographies so that the bottom of the last one
% is flush with the other column.
%\enlargethispage{-5in}

% that's all folks
\end{document}